\documentclass[sigconf]{acmart}

\usepackage{amsmath}
\usepackage{multirow}
\usepackage{booktabs}
\usepackage{balance}

\AtBeginDocument{%
  \providecommand\BibTeX{{%
    \normalfont B\kern-0.5em{\scshape i\kern-0.25em b}\kern-0.8em\TeX}}}

\copyrightyear{2020} 
\acmYear{2020} 
\setcopyright{acmcopyright}
\acmConference[MM '20]{Proceedings of the 28th ACM International Conference on Multimedia}{October 12--16, 2020}{Seattle, WA, USA} 
\acmBooktitle{Proceedings of the 28th ACM International Conference on Multimedia (MM '20), October 12--16, 2020, Seattle, WA, USA} 
\acmPrice{15.00} 
\acmDOI{10.1145/3394171.3413644} 
\acmISBN{978-1-4503-7988-5/20/10}

\acmSubmissionID{1394}

\begin{document}
\fancyhead{}

\title{Learning Global Structure Consistency\\ for Robust Object Tracking}

\author{Bi Li}
\authornote{This work is done while Bi Li is interning at Baidu Inc.}
\affiliation{%
  \institution{School of Electronic Information and Communications, Huazhong University of Science and Technology}
  }
\email{libi@hust.edu.cn}

\author{Chengquan Zhang}
\affiliation{%
  \institution{Department of Computer Vision Technology (VIS), Baidu Inc.}
}
\email{zhangchengquan@baidu.com}

\author{Zhibin Hong}
\affiliation{%
  \institution{Department of Computer Vision Technology (VIS), Baidu Inc.}
}
\email{hongzhibin@baidu.com}

\author{Xu Tang}
\affiliation{%
  \institution{Department of Computer Vision Technology (VIS), Baidu Inc.}
}
\email{tangxu02@baidu.com}

\author{Jingtuo Liu}
\affiliation{%
  \institution{Department of Computer Vision Technology (VIS), Baidu Inc.}
}
\email{liujingtuo@baidu.com}

\author{Junyu Han}
\affiliation{%
  \institution{Department of Computer Vision Technology (VIS), Baidu Inc.}
}
\email{hanjunyu@baidu.com}

\author{Errui Ding}
\affiliation{%
  \institution{Department of Computer Vision Technology (VIS), Baidu Inc.}
}
\email{dingerrui@baidu.com}

\author{Wenyu Liu}
\authornote{Corresponding author.}
\affiliation{%
  \institution{School of Electronic Information and Communications, Huazhong University of Science and Technology}
}
\email{liuwy@hust.edu.cn}

\renewcommand{\shortauthors}{Bi and Chengquan, et al.}

\begin{abstract}
  Fast appearance variations and the distractions of similar objects are two of the most challenging problems in visual object tracking. 
  Unlike many existing trackers that focus on modeling only the target, in this work, we consider the \emph{transient variations of the whole scene}. 
  The key insight is that the object correspondence and spatial layout of the whole scene are consistent (i.e., global structure consistency) in consecutive frames which helps to disambiguate the target from distractors. Moreover, modeling transient variations enables to localize the target under fast variations. 
  Specifically, we propose an effective and efficient short-term model that learns to exploit the global structure consistency in a short time and thus can handle fast variations and distractors.
  Since short-term modeling falls short of handling occlusion and out of the views, we adopt the long-short term paradigm and use a long-term model that corrects the short-term model when it drifts away from the target or the target is not present. These two components are carefully combined to achieve the balance of stability and plasticity during tracking.
  We empirically verify that the proposed tracker can tackle the two challenging scenarios and validate it on large scale benchmarks. Remarkably, our tracker improves state-of-the-art-performance on VOT2018 from 0.440 to 0.460, GOT-10k from 0.611 to 0.640, and NFS from 0.619 to 0.629.
\end{abstract}



\maketitle

\section{Introduction}

\begin{figure}[h]
  \centering
  \includegraphics[width=0.7\linewidth]{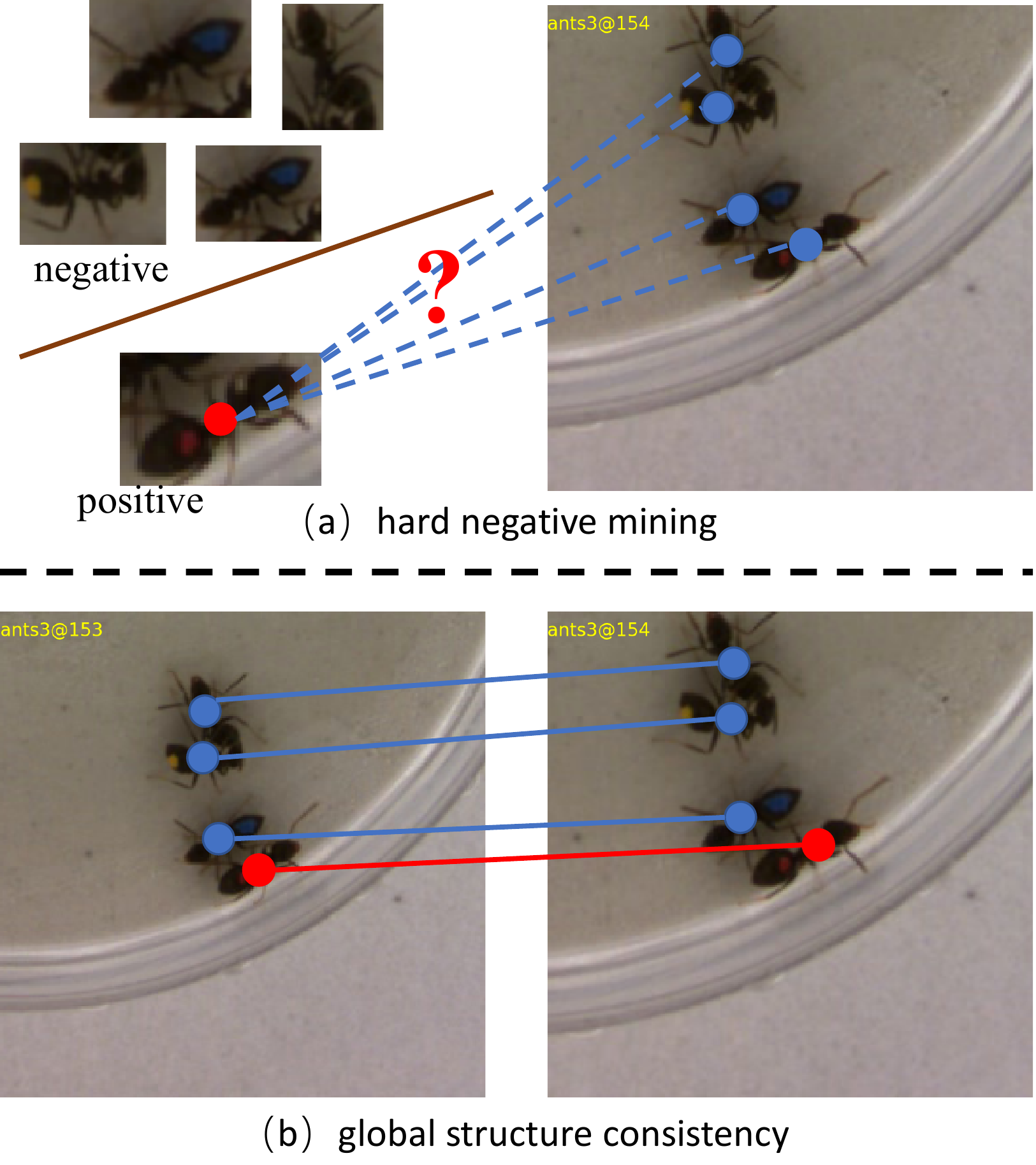}
  \caption{To find the target in the current frame, (a) methods based on hard negative mining struggle to differentiate the target from distractors since they all belong to the ``ant'' category and share similar appearance (b) by considering the global structure (object correspondence and spatial layout) of the scene which is consistent in two consecutive frames, the target is easy to find. \label{fig:gsc}}
  \Description{The global structure consistency.}
\end{figure}

Object tracking aims to localize the target object in each frame of a video given only the target position in the first frame. Typically, the target position is encoded as a rectangle. Object tracking is unique in the sense that it is instance-centered. Instead of localizing all instances of some specified categories like object detection, it focuses on one \emph{specific} instance from arbitrary categories. The target object to be tracked is only known when the first frame of the video is given. This demands a specific target model to be learned on-the-fly during tracking, which is challenging due to limited supervision.

Recent years, thanks to the representation capability of deep neural networks, 
trackers\cite{bertinetto_fully-convolutional_2016, li_high_2018, zhang_deeper_2019, li_siamrpn_2019, bhat_unveiling_2018, danelljan_atom_2019, bhat_learning_2019} can now localize targets under large variations of poses, sizes and illuminations, etc. 
However, it remains difficult to track the target under fast variations (rare pose during object transformation, motion blur, etc.) and visually similar distractors. 

One reason behind the difficulty for localizing target under fast variations is that most trackers adopt a single target model and a conservative update strategy to avoid drifting.
 For example, \cite{nam_learning_2016, danelljan_eco_2017, danelljan_atom_2019,bhat_learning_2019} update the target model every 10 frames with few iterations. Siamese trackers\cite{bertinetto_fully-convolutional_2016, li_high_2018, zhang_deeper_2019, li_siamrpn_2019} do not update the target model at all. 
 When the tracker has only a single target model with a conservative update strategy, transient details in the present are not absorbed into the single target model as quickly as possible.
 However, to localize the target under fast variations, it requires modeling these details which results in the tracking failure.
 
 As for distractors, most existing trackers tackle the problem by using either hard negative learning\cite{nam_learning_2016,zhu2018distractor} or cosine window\cite{bertinetto_fully-convolutional_2016} to suppress objects in the background. However, since distractors are visually similar and often among the same semantic category as the target, appearance-based hard negative learning remains difficult. On the other hand, the cosine window cannot handle fast-moving targets that are far away from the center of the search image. 
 
 To tackle the fast variation and distractor problem, we focus our attention on the short-term variation of consecutive frames. Models that can capture short-term variations are naturally suited to track the target in fast variation. Moreover, we show it can also help with the distractors.
 Our key observation is that, under short-term variation, the global structure (including object correspondence and spatial layout) of the whole scene is consistent, which is particularly helpful for discriminating distractors. For an illustrative explanation of the global structure consistency (GSC), please refer to Figure~\ref{fig:gsc}. 

In this work, we design a neural network that can models short-term variations such that it can capture targets under fast variations. Crucially, it is also able to exploit the global structure consistency of consecutive frames to suppress distractors. To quickly adapt to transient variations without risking model drift, we adopt the long-short term design which contains a long-term component to model the "whole picture" of the target and a short-term component to model the short-term variations. By sharing the backbone features, the proposed long-short term tracker is as efficient as trackers with only one target model. We extensively evaluate the proposed tracker on large scale tracking benchmarks (VOT2018\cite{kristan_sixth_2018}, OTB-100\cite{wu_object_2015}, GOT-10k\cite{huang_got-10k_2019}, TrackingNet\cite{muller_trackingnet_2018}, NFS\cite{kiani_galoogahi_need_2017}, LaSOT\cite{fan_lasot_2019}) and demonstrates state-of-the-art performance.

In summary, the contributions are threefold:
\begin{enumerate}
  \item We propose an efficient and effective short-term model that learns the global structure consistency and can handle challenging scenarios including fast variations and distractors.
  \item We propose an effective method to combine the long-term and short-term models such that the tracker can quickly adapt to target variations without model drift. 
  \item We evaluate the proposed tracker extensively on large scale benchmarks and demonstrate state-of-the-art performance.
\end{enumerate}

\begin{figure*}[h]
  \centering
  \includegraphics[width=0.86\linewidth]{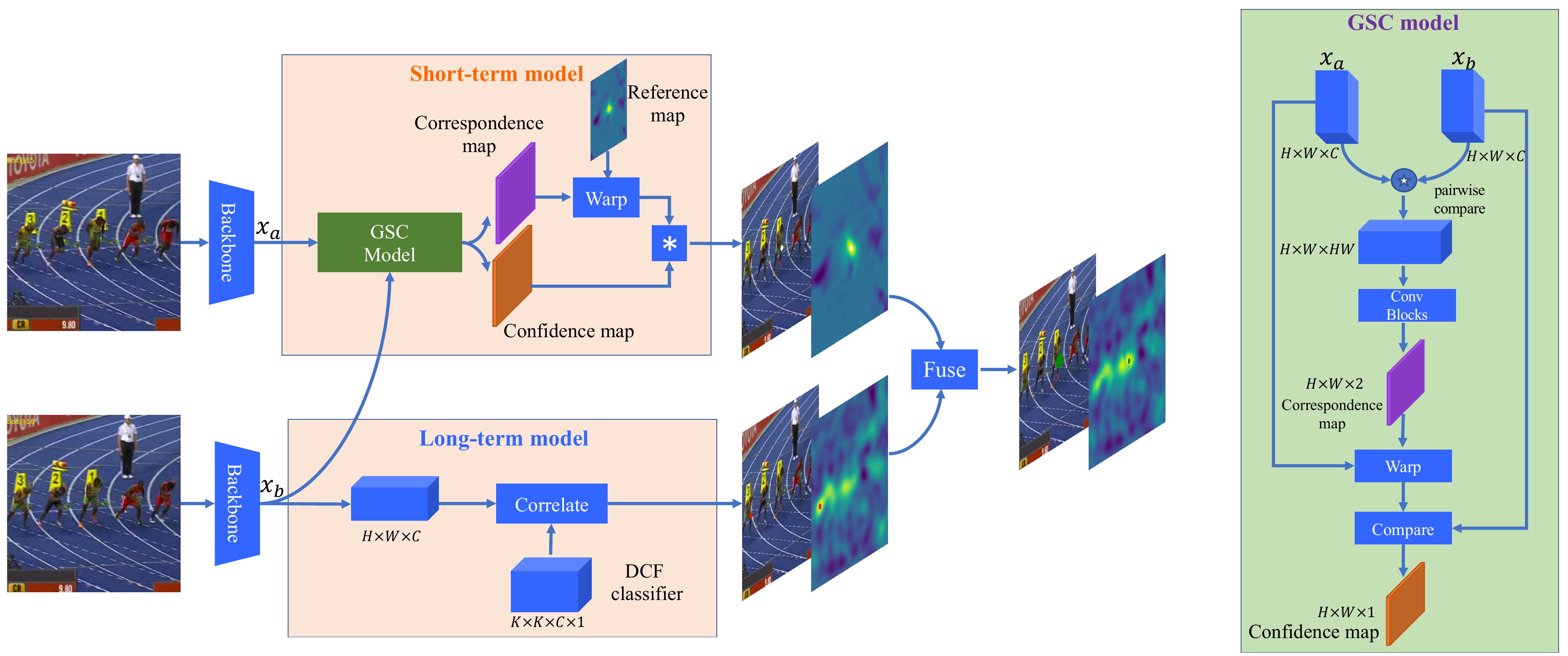}
  \caption{The framework of the proposed method. Our tracker contains a short-term and a long-term model. During tracking, at one step, the short-term model takes as input the last and current image and outputs the response map of the current image by establishing the correspondence between the two images and then warp the reference map at the last step. The long-term model shares the backbone with the short-term model and is based on a discriminative correlation filter(DCF) to obtain the response map of the current image. These two response maps are fused to obtain the final response. \label{fig:framework}}
\end{figure*}

\section{Related Work}
\subsection{Long-short term Trackers}
Stability and plasticity dilemma\cite{grossberg_competitive_1987} is a long-standing problem in object tracking, i.e., it should be stable with respect to noisy supervision while being flexible to quickly adapt to target variations. To tackle this problem, various trackers have been proposed in a two-component architecture: a long-term component for stability and a short-term component for plasticity. Generally, the long-term component is conservatively updated to keep stable even under occlusion and out of the views. In contrast, the short-term component is aggressively updated to quickly adapt to target variations and suppress distractors. The key design choices of these trackers are how to model the long-term and short-term variations and how these two components are combined.

TLD\cite{kalal2011tracking} decomposes the tracking system into a tracker and a detector, where the tracker is based on the pyramid Lucas-Kanade tracker\cite{Bouguet1999PyramidalIO}, and the detector is a cascaded classifier. Each component outputs the estimated target bounding box which are then combined based on the conservative similarity.
Inspired by the Atkinson-Shiffrin Memory Model, Muster\cite{hong2015multi} proposes a long-term memory and a short-term memory for object tracking. Correlation filters are used for short-term memory and keypoints matching for long-term memory. The two bounding boxes outputted by the components are integrated based on the tracking state. PTAV\cite{fan2017parallel} uses a Siamese network for verification and a correlation filter for real-time tracking. The verifier is invoked every 10 frames to verify if the tracking results are correct. Otherwise, the tracker will be reset by the verifier. 
Instead of using two different models for the long- and short-term component, LCT\cite{ma2018adaptive} uses correlation filters for both components. The tracker outputs the target position predicted by the short-term component unless the confidence scored by the long-term component is lower than a threshold. THOR\cite{Sauer2019BMVC} uses templates for both components. The long-term module is only updated when the gram determinant, which indicates the templates diversity, is increased. The short-term module is constantly updated with the latest target template. In contrary to combine bounding boxes output by two components, THOR fuses the activation maps of all templates and then gets the bounding box.

Our tracker follows the heterogeneous design of \cite{kalal2011tracking,hong2015multi,fan2017parallel} and uses two different models for short-term and long-term modeling. However, the backbone features are shared in these two components and thus the efficiency is not compromised. We follow \cite{Sauer2019BMVC} and fuse the response map which provides fine-grained control between stability and plasticity.

\subsection{Background Modeling Trackers}
Global structure consistency can be seen as one form of background or context modeling. Dinh et al\cite{dinh2011context} exploits the context by modeling the supporters and distracters. Xiao et al\cite{xiao2016distractor} detects distractors and model the context information of the target in terms of the distractor positions. All these methods require explicitly detecting background objects. Contrarily, our method implicitly models the background and uses neural networks to learn global structure consistency.

Lucas-Kanade trackers\cite{lucas_iterative_1981, shi_good_1994} establish the correspondence of sampled points (including both target and background) between two adjacent frames and resemble our short-term model. However, these methods assume constant brightness which is not always available. Instead, our method learns to estimate the correspondence using large number of offline data and can handle challenging scenarios. Recently, using convolutional neural networks to learn dense correspondence between two images have been actively explored\cite{dosovitskiy_flownet_2015,ilg_flownet_2017, sun_pwc-net_2018,melekhov_dgc-net_2019}. However, pretrained optical flow models are not optimal for short-term modeling for several reasons. First of all, since gathering dense correspondence labels in real data is hard, optical flow models are typically trained using synthetic data. Second, optical flow models lack the correspondence confidence which is important for combining the short- and long-term components. Finally, optical flow models are computationally heavy. Our method overcomes all these limitations and demonstrates to be efficient and effective.

\section{Methods}
Our tracker follows the long-short term design which contains a long-term and a short-term component. The long-term component is responsible for modeling the overall characteristics of the target in a long time and supposed to be reliable even after occlusion and out of the views. Specifically, a discriminative correlation filter is used which is conservatively updated to keep stable. The short-term component aims to capture the transient characteristics of the target and can localize the target under fast variations and distractors. We exploit the global structure consistency in the short-term model by establishing a correspondence map between two adjacent frames and then warp the reference map based on the correspondence to get the short-term response. These two components are then carefully fused to maintain a good balance between stability and plasticity. The whole framework is illustrated in Figure~\ref{fig:framework}. 

In the following, we exploit the global structure consistency in the short-term component (Sec~\ref{sec:short-term}) to achieve robust tracking. We briefly introduce the long-term component (Sec~\ref{sec:long-term}) and describe how the two components are combined (Sec~\ref{sec:combine}). Implementation details are also included (Sec~\ref{sec:detail}). 

\subsection{Short-term Modeling} \label{sec:short-term}

\textbf{Global Structure Consistency.}
One of the most important distinctions between long-term and short-term variation is whether the temporal consistency exists. Temporal consistency stems from the fact that the world is continuous on a macro scale. Therefore, the difference between image frames capturing the world in a short period is small. Most trackers have already exploited the temporal consistency to smoothen the target trajectory. For example, it is common practice to penalize large displacement of the target centers between two adjacent frames. 

While it has been successfully applied in object tracking, most trackers focus on the temporal consistency of only the \emph{target} and completely ignores the \emph{background}. This is understandable because the tracking task requires only the positions of the target and modeling the large area of background adds computational burdens for practical usage. Nevertheless, as we will show, by taking the background into consideration, it helps to robustly localize the target under challenging scenarios, especially under distractors, and the additional computation is marginal with good design. 

Crucially, we observe that in a short time, the object correspondence and spatial layout in the scene are consistent. 
We dub this phenomenon \emph{global structure consistency} (GSC). 
We use \emph{global} to emphasize that it includes both the target and objects in the background. Intuitively, if the background objects are falsely recognized as the target, the global structure consistency will be broken. This helps to disambiguate the target from similar distractors. For a visual explanation of the global structure consistency, please refer to Figure~\ref{fig:gsc}. We demonstrate how to exploit the GSC in the following sections.

\textbf{Formulation of the short-term problem.} We first formulate the short-term problem that we aim to solve: Given two adjacent frames $[f_a, f_b]$ and the target position $p_a$ in frame $f_a$, estimate the target position $p_b$ in frame $f_b$. The frame could be the full image or the cropped and scaled search image. In this work, we use search image crops $f \in R^{H_f\times W_f \times 3}$. The target position $p \in \mathbb{R}^4$ specifies the $[$target center, height, width$]$. Generally, to solve the short-term problem above, the model should capture the transient variations of the targets.

\textbf{Our approach with GSC.} Both frames $[f_a, f_b]$ are first embedded by a backbone network $\varphi(\cdot)$ to get the features $x = \varphi(f) \in \mathbb{R}^{H\times W\times C}$. After extracting the features of the two frames $x_a, x_b$, we then use all $1\times 1$ features from $x_a$ and cross-correlates them with $x_b$, leading to a response map $r \in \mathbb{R}^{H \times W \times HW}$. This operation can be readily implemented using reshape and matrix multiplication as below:

\begin{align}
  \label{eqn:cross}
  \begin{split}
    X_a &= \text{reshape}(x_a, (HW, C))\\
    X_b &= \text{reshape}(x_b, (HW, C))\\
    R &= X_bX_a^T\\
    r &= \text{reshape}(R, (H, W, HW))
  \end{split}
\end{align}

In this way, all correspondence between the two frames are taken into consideration. Moreover, the spatial layout of the objects are implicitly encoded in the response maps. We then use several convolution blocks (Conv + BN + ReLU) to project the response map $r$ into the correspondence map $c = \text{ConvBlocks}(r) \in \mathbb{R}^{H \times W \times 2}$. In the correspondence map, each element $(y_a, x_a) = c[y_b, x_b] \in \mathbb{R}^2$ tells the corresponding position $(x_a, y_a)$ in $f_a$ of $(x_b, y_b)$ in $f_b$. Typically, the image size $[H_f, W_f]$ is multiple times larger than the size of their feature map $[H, W]$. In this case, we scale the correspondence map (which is of the same size as the feature) up into the image size using bilinear upsampling. This is the default method for handling size mismatch unless otherwise specified. Given the target center in frame $f_a$ and the correspondence map $c$, we can readily find the target center in frame $f_b$. The height and width of the target will use those in frame $f_a$.

\textbf{Confidence estimation.} In the method above, only the correspondence map $c$ is regressed, and it is not clear how confident the correspondences are. It is very likely that some of them in the correspondence map $c$ are incorrect. Therefore, the confidence score is important for combining long-term and short-term components. The correspondence is only reliable when the confidence is high.

To estimate the confidence of each correspondence in the correspondence map $c$, we warp the feature $x_a$ based on the correspondence map $c$ to generate $\tilde{x}_a = \text{warp}(x_a, c) \in \mathbb{R}^{H\times W\times C}$. Now each element in the warped feature $\tilde{x}_a$ is  aligned with elements in feature $x_b$. We then compare the similarity between the aligned elements, the confidence of correspondence is positively correlated to the similarity. Specifically, the warped feature is elementwise multiplied with $x_b$ and summed in the channel dimension which is then followed by a ReLU function to suppress noise. The operations are summarized in below:

\begin{align}
  \label{eqn:confidence}
  \begin{split}
    \tilde{x}_a &= \text{warp}(x_a, c)\\
    s &= \text{relu}(\sum_{\text{channel}}(\tilde{x}_a \odot x_b))
  \end{split}
\end{align}

Note that instead of directly regressing the confidence map based on two features, $s = \text{EstimateConf}(x_a, x_b)$, our approach estimates the confidence map depending on the correspondence map, $s = \text{EstimateConf}(x_a, x_b, c)$. This is beneficial since now we can use any correspondence map with a known confidence map for training. Otherwise, it will be difficult to design the groundtruth label for the confidence map without overfitting.

\begin{figure}[t]
  \centering
  \includegraphics[width=1.0\linewidth]{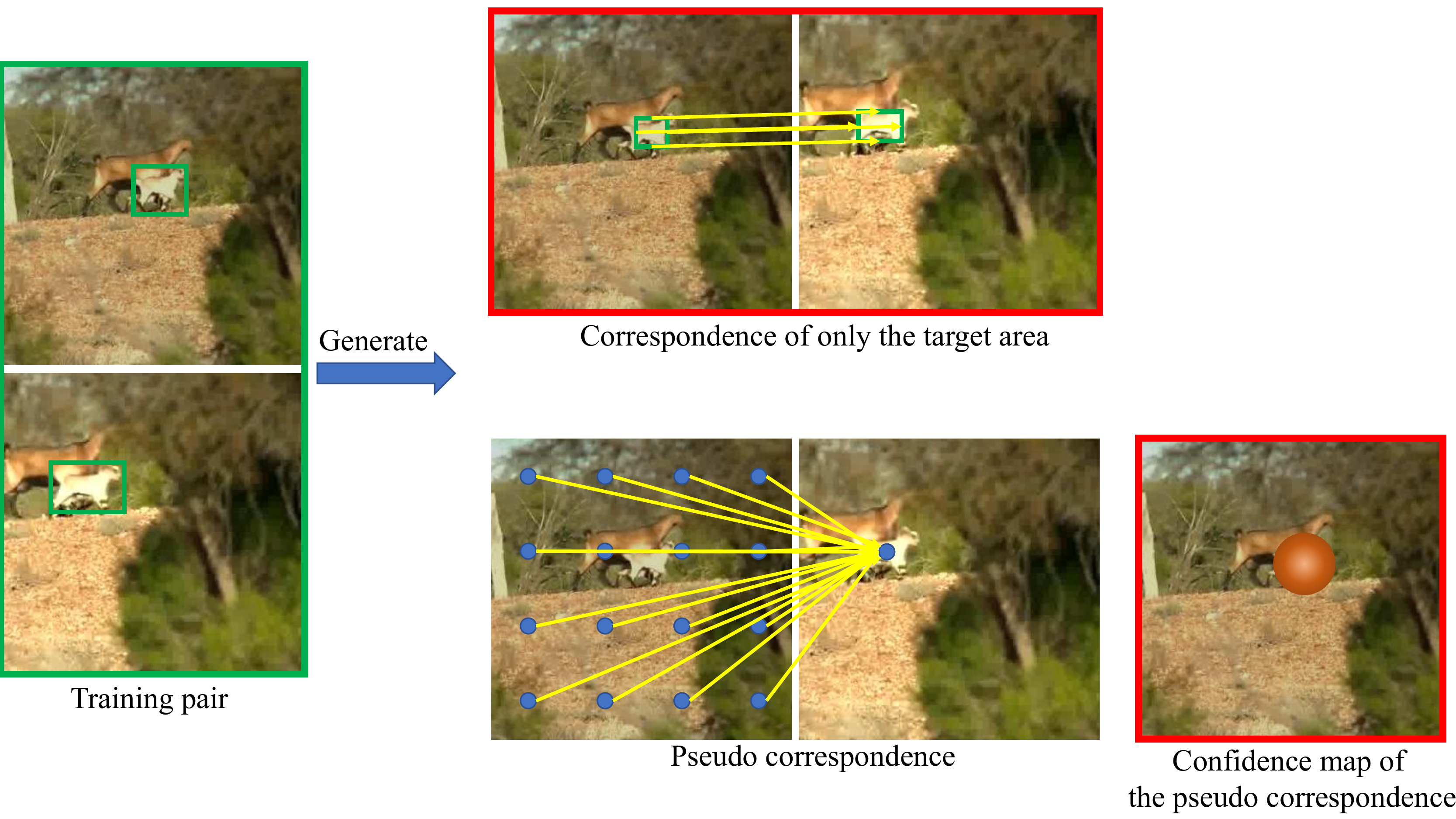}
  \caption{Given a pair of training images with bounding box annotations, we generate the correspondence map considering only the target areas and the confidence map of a pseudo correspondence map. Please refer to Sec~\ref{sec:train_label} \textit{Training label and loss} for details. \label{fig:labels}}
\end{figure}

\textbf{Training label and loss.} \label{sec:train_label}
During offline training, given the target positions $[p_a, p_b]$ in two adjacent frames, we now describe how to generate the training labels including the correspondence map $c$ and the confidence map $s$. For a visual illustration of the training labels, please refer to Figure~\ref{fig:labels}.

For the correspondence map, we first estimate the affine transformation of the two frames based on the two target positions and then use the affine transform to generate the correspondence map. It is clear that only the target area of the correspondence is correct. Therefore, the loss at each correspondence is weighted by a mask where the target area is 1 and the background area is 0. To avoid including background in the target area, we shrink the target bounding box by a factor of 2 as the target area. Even though the correspondence map is supervised only in the target area during training, since the approach is target-agnostic during inference, it can still predict correspondences of all objects in the image.

To get the groundtruth of the confidence map, we first need a correspondence map. It seems natural to use the above groundtruth correspondence map. However, this is problematic since it only has correct correspondences in the groundtruth and the confidence will always be 1. The confidence branch will overfit to output all 1s. To mitigate this problem, we  generate a pseudo correspondence map $\tilde{c} \in \mathbb{R}^{H\times W \times 2}$. Each element $\tilde{c}[y, x] = (y_c, x_c)$ where $(y_c, x_c)$ is the target object center in $f_a$. In other words, all positions in $f_b$ corresponds to the target center in $f_a$ in the pseudo correspondence map. The confidence map of the pseudo  correspondence is a gaussian function with respect to the $L_2$ distance between the current position and the target center in $f_b$. During training, we use the pseudo  correspondence map to estimate the confidence which is supervised by the groundtruth confidence map.

Both predictions of the correspondence map and the confidence map are compared to their groundtruth using the $L_1$ loss. These two losses are then summed as the final loss for training.

\subsection{Long-term Modeling} \label{sec:long-term}
For the long-term component, we follow the ATOM tracker\cite{danelljan_atom_2019} and use a discriminative correlation filter (DCF) for localizing the target center and an IoU-Net\cite{jiang_acquisition_2018} for refining target position. Compared to ATOM, we improve the long-term component by 1) using a deeper backbone (ResNet-50 instead of ResNet-18) and 2) fixing the BatchNorm layers of the backbone during offline training. As shown in the ablation study (Section \ref{sec:ablation}), these two improvements are important for the overall tracking performance. During tracking, we use the same hyperparameters as ATOM.

Basically, the DCF classifier is initialized in the first frame using augmented training samples (translation, rotation, etc.). The classifier is updated under two cases: 1) every 10 frames or 2) distractors are present. Moreover, it is only updated when the classifier is confident that the target is present (response higher than a threshold). This ability makes it reliable under occlusions and out-of-views. Gaussian-Newton method is used to learn the classifier. With the update mechanism, the DCF classifier already has capabilities to model target variations. However, it cannot quickly learn variations within 10 frames, and empirically it still has difficulty learning to disambiguate the distractors. On the other hand, our short-term component is suited to these scenarios.

Given a frame $f_t$ at time $t$, the same backbone $\varphi(\cdot)$ of the short-term component is used to get the features $x_t$. Sharing the backbone network is important for reducing computations. The DCF classifier is then applied to the feature map to get the long-term response map $r^l \in \mathbb{R}^{H\times W\times 1}$ where the position with the highest score is taken as the target center. The width and height of the target are the same as those in the last frame.

The IoU-Net is a neural network that takes as input the feature $x$ and target position $p$ and outputs the estimated intersection over union (IoU) between $p$ and groundtruth target position. Importantly, the estimated IoU of the IoU-Net is differentiable with respect to the target position $p$ and thus stochastic gradient descent(SGD) is used to maximize the estimated IoU by adjusting the target position $p$. In ATOM, IoU-Net is used to refine the target bounding box estimated by the DCF classifier.

\subsection{Long/short-term combination} \label{sec:combine}
Given the long-term and the short-term component, it is important to fuse them by their reliability. In principle, the short-term component is helpful if two conditions are satisfied: a) the estimated target positions $p_a$ in the last frame is reliable; b) the estimation of the short-term component in the current frame is reliable. Otherwise, the tracker should be inclined to trust the long-term component. 

At time $t$, given the frame $f_t$, the DCF classifier of the long-term component is used to compute a long-term response map  $r^l_{t} \in \mathbb{R}^{H\times W\times 1}$ which indicates the likelihood of the target center in different positions. The short-term component takes as input two adjacent frames $[f_{t-1}, f_{t}]$ and outputs the correspondence map $c$ and the confidence map $s$. To combine with the long-term response $r^l_t$, we generate a short-term response map based on the correspondence map $c$ and the confidence map $s$, which should also indicate the center of the target. The procedures are described below.

Given the target position $p_{t-1}$ at time $t-1$, we first generate an reference map $m$ by centering a gaussian on the target center and then weights the gaussian map by the long-term response $r^l_{t-1}$ at time $t-1$. Intuitively, when the $r^l_{t-1}$ is high, the tracker is confident that the estimated target position in the last frame is correct and vice versa. Then, we warp the reference response $m$ using the correspondence map $c$ to get the transferred response map. We then weight the transferred response map with the clipped confidence map $T_s + \text{relu}(s - T_s)$ to suppress unreliable responses where $T_s$ is a hyperparameter that controls the low confidence threshold. In summary, the short-term response map is:

\begin{align}
  \label{eqn:short}
  \begin{split}
    m &= r^l_{t-1} \ast \text{Gaussian}(p_{t-1})\\
    r^s_{t} &= (T_s + \text{relu}(s - T_s)) \ast \text{warp}(m, c)
  \end{split}
\end{align}

Given the long-term response $r^l_{t}$ and the short-term response $r^s_t$, we further weight the short-term response by the local maximum mask $k$ of the long-term response. This is because we find that the long-term response is more accurate in terms of center localization while short-term response helps to disambiguate distractors. The local maximum mask can be readily implemented using the max pooling and equal operation. Finally, they are linearly weighted to get the final response:

\begin{align}
  \label{eqn:combine}
  \begin{split}
    k_t &= (\text{maxpool}(r^l_t, K) == r^l_t)\\
    r^{ls}_t &= r^l_t + W * k_t * r^s_t 
  \end{split}
\end{align}
where $K$ and $W$ are hyperparameters, $K$ is the kernel size of the max pooling operation, $W$ is the fusion weight. Note that we use constant weight 1.0 for the long-term response such that the response map reduces to the long-term response when the short-term response is not reliable. The position with the highest response is taken as the target center, and we then use the IoU-Net to get the final target position. 

In long videos, the target is sometimes completely occluded or out of the view. To handle these cases, we use the score of the long-term component to estimate the target state. If the highest score in the long-term response map is lower than a threshold, the target is lost and we set the weight of the short-term component to zeros, $W = 0$. Even after the target reappears, since it is not clear whether the re-detected object is truly the target, we decay the fusion weight by a factor $D$ thereafter.

\subsection{Implementation Details} \label{sec:detail}
\textbf{Features}. The backbone is shared for both the long-term and the short-term component. Therefore, the extra computation of the short-term component is marginal. For the short-term component, we use the Layer3 features from ResNet-50\cite{He2016DeepRL} and add a bottleneck layer to downsample the feature size to 512. For the long-term component, we use the Layer3 features for the DCF classifier, Layer3 and Layer4 features for the IoU-Net. The backbone is pretrained with ImageNet\cite{russakovsky_imagenet_2015}. During training the short-term or the long-term component, the backbone is fixed. 

\textbf{Training}. ImageNet VID\cite{russakovsky_imagenet_2015}, LaSOT\cite{fan_lasot_2019}, COCO\cite{lin_microsoft_2014}, GOT-10k\cite{huang_got-10k_2019} and YouTube-BoundingBox\cite{real_youtube-boundingboxes_2017} datasets are used for training. All datasets are sampled with equal probability. Videos that may be in the evaluation benchmarks are all carefully removed. We first train the long-term component and then the short-term component. For the long-term component, all training configurations (data sampling, batch size, learning rate, optimizer, epoch, etc.) are the same as ATOM except that each epoch has 6000 iterations. Moreover, we fix the BatchNorm\cite{ioffe_batch_2015} of the backbone during training which is unintentionally updated in ATOM.

For the short-term component, instead of sampling two adjacent frames from a video, we sample frames that are within a short range ($\le 10$). In this way, the diversity of training samples are increased and the trained model can better handle various frame rate. It is trained for 50 epochs, each epoch with 5000 iterations. ADAM\cite{kingma_adam_2015} is used to optimize the short-term model with initial learning rate 1e-2 which is multiplied by 0.1 every 15 epochs.

\textbf{Hyperparameters}. During tracking, we use the same sets of hyperparameters for all benchmarks except on VOT. This is due to the evaluation protocol of VOT has a tracker re-initialization mechanism. For the long-term component, we use the same hyperparameters as the ones in ATOM. For the short-term component, the hyperparameters are manually tuned based on our experience. Specifically, the reliable threshold $T_s = 0.4$, the kernel size of the local maximum operation $K = 8$, the long-term and short-term fusion weight $W = 0.6$, the decay factor of the fusion weight $D = 0.6$.

\begin{figure*}[h]
  \centering
  \includegraphics[width=0.90\linewidth]{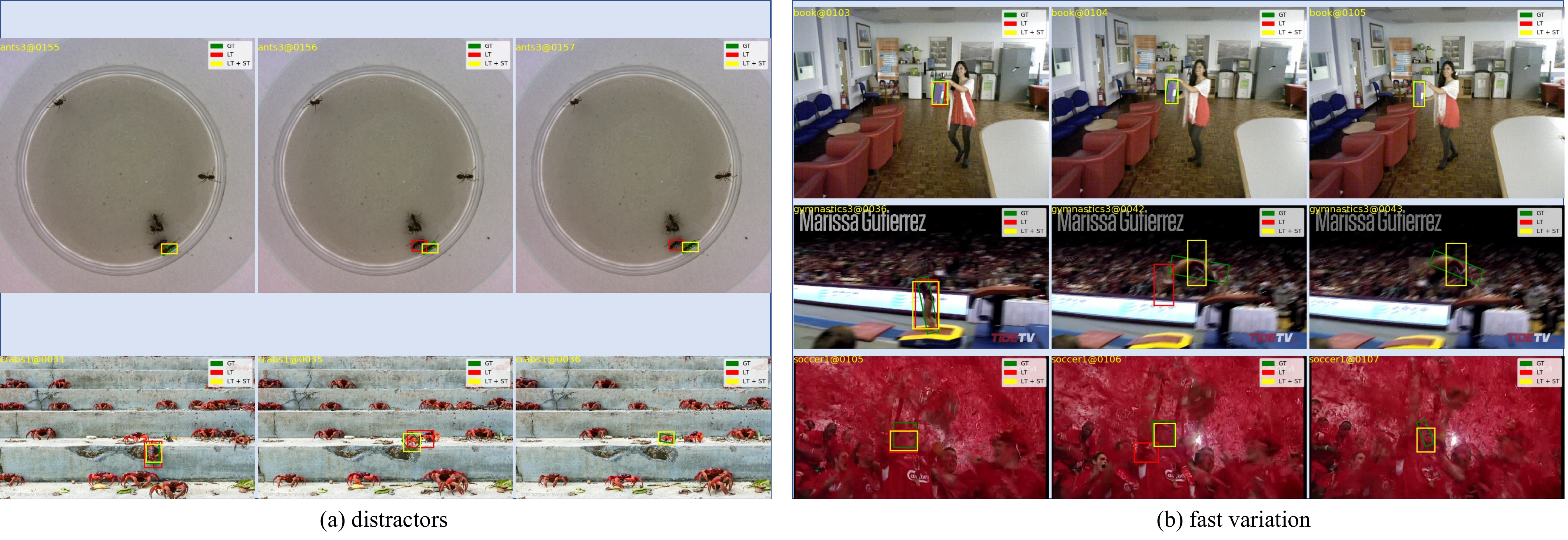}
  \caption{Qualitatively comparing the results with/without the short-term component. Videos are selected from VOT2018. The groundtruth in \textcolor{green}{green} box. The tracking results using only the long-term component are shown in \textcolor{red}{red} box. The results using both long-term and short-term components are shown in \textcolor{yellow}{yellow} box. The box is not present if the overlap between the prediction and the groundtruth is zero. As can be seen, by adopting the short term component with GSC modeling, our tracker can effectively handle two challenging scenarios including fast variation and distractors. Best viewed in digital device with zooming capability. \label{fig:qualitative}}
\end{figure*}

\section{Experiments}

\subsection{Ablation Study} \label{sec:ablation}
In this section, we perform an ablation study of the proposed tracker on the VOT2018\cite{kristan_sixth_2018} dataset. Since our tracker is stochastic, all experiments are repeated 15 times following the standard evaluation protocol of VOT2018. All experiments share the same hyperparameters except the ones to inspect. The results are summarized in Table~\ref{tab:ablation}.

\begin{table}[t]
  \centering
  \begin{tabular}{c | c  c  c | c | c c}
    \hline
    \multirow{2}{*}{Backbone} & \multicolumn{3}{c|}{short-term} & long-term  &  \multirow{2}{*}{EAO} & \multirow{2}{*}{FPS}\\ \cline{2-5}
     & OF & GSC & Conf.                    &  Fix BN    &      \\
    \hline
    Res-18  &  &               &               &              & 0.399  & 16 \\
    Res-18  &  &               &               &   \checkmark & 0.409  & 16 \\
    Res-50  &  &               &               &   \checkmark & 0.441  & 14 \\
    \hline
    Res-50  & \checkmark & & \checkmark & \checkmark & 0.450 & 7 \\
    \hline
    Res-50  &  &  \checkmark   &               & \checkmark & 0.442 & 13 \\
    Res-50  &  &               &   \checkmark  & \checkmark & 0.452 & 13 \\
    \hline
    Res-18  &  &  \checkmark   &   \checkmark  & \checkmark & 0.431 & 15\\
    Res-50  &  &  \checkmark   &   \checkmark  & \checkmark & 0.460 & 13 \\
    \hline
  \end{tabular}
    \caption{Ablation studies on the VOT2018 dataset. Different backbones are investigated. The impacts of the optical flow (OF), global structure consistency (GSC) and the confidence (Conf.) estimation of the short-term component are shown. The impacts of whether fixing the BatchNorm (BN) of the backbone during training is shown. Please refer to Section~\ref{sec:ablation} for detailed analysis.}
  \label{tab:ablation}
\end{table}

\textbf{Impacts of backbone features.} It is widely acknowledged that backbones are vital to the performance of visual tasks. We investigate the impact of using deeper backbones. By changing the backbone from ResNet-18 to ResNet-50, the performance is largely improved. When only using the long-term component, the EAO improves from 0.409 to 0.441. When using both components, the EAO improves from 0.431 to 0.460.

Moreover, the BatchNorm layers of the backbone in ATOM are unintentionally updated during training, this can be problematic for our tracker since both long- and short-term components share the same backbone. We fix the BatchNorm and it brings another 1\% absolute gain, from 0.399 to 0.409. 

\textbf{Impacts of short-term component}. With the short-term modeling capability, the EAO improves from 0.441 to 0.460. It is noteworthy that 0.441 is already a state-of-the-art result on the VOT2018 benchmark. Nevertheless, our short-term component can still largely improve tracking performance. This validates the importance of short-term modeling. Some qualitative results are shown in Figure~\ref{fig:qualitative}. Notably, the short-term component helps when the target undergoes fast variations or distractors. Moreover, the extra computation of the short-term component is marginal. The tracking speed slightly decreases from 14 FPS to 13 FPS.

\textbf{Impacts of short-term confidence}. The confidence of the short-term component is important for combining the short-term and long-term components. We validate this by removing the confidence estimation module and use a constant confidence map with value $1$s instead. The performance degrades from 0.460 to 0.442. Without the mechanism to filter out unreliable short-term results, the short-term component only marginally improves the performance from 0.441 to 0.442. 

\textbf{Impacts of global structure consistency}. To demonstrate the effectiveness of modeling the GSC, we design a simple baseline without GSC for comparison. After extracting the features of both frames $x_a, x_b$, we pool the target feature at position $p_a$ from $x_a$ using ROI pooling $\bar{x} = \text{roipool}(x_a, p_a) \in \mathbb{R}^{H_o \times W_o \times C}$ where $H_o$ and $W_o$ are the size of the pooled feature. Specifically, we use $H_o = W_o = 3$ in the experiments. The pooled target feature is then cross-correlated with the $x_b$ to get the response map $r = x_b \star \bar{x} \in \mathbb{R}^{H \times W \times 1}$. The position with the highest response is taken as the estimated center of the target. The height and width of the target is the same as those in frame $f_a$. Notice that in this baseline, the GSC is not modeled since only the target area in frame $f_a$ is used for estimating the target position in frame $f_b$. Compared to our method, the performance of the baseline degrades from 0.460 to 0.452. Without modeling the global structure consistency, the tracker has difficulties in handling distractors.

\textbf{Comparison with optical flow model}. We compare our short-term model with optical flow models. We use pretrained PWCNet\cite{sun_pwc-net_2018} for computing the optical flow which is used as the correspondence map in our framework. Since the PWCNet does not have a confidence map, we use the forward-backward error to indicate the confidence. By using the PWCNet as the short-term model, the performance improves from 0.441 to 0.450. However, it is worse than our GSC based model which achieves 0.460. Moreover, since the backbone features are not shared for the optical flow model, it runs much slower, 7 fps compared to 13 fps.

\subsection{State of The Art Comparison}
In this section, we extensively compare our proposed tracker with the state-of-the-art approaches on large scale benchmarks including VOT2018, OTB-100, NFS, LaSOT, GOT-10k, and TrackingNet. To further demonstrate the benefits of adopting the long-short term design, we also show the results of our tracker with only long-term component, which is denoted as Ours-LT.

\textbf{VOT2018}\cite{kristan_sixth_2018}: VOT2018 dataset consists of 60 challenging videos and focuses on short-term tracking (no redetection is required and resets the tracker once it fails). In the short-term tracking scenario, the short-term component is more helpful since the estimated target position in the last frame is typically reliable. Otherwise, the tracker would have been reset following the evaluation protocol. Trackers are evaluated based on the measures robustness, accuracy and expected average overlap (EAO). Robustness indicates the number of tracking failures which is the lower the better. Accuracy is the average overlap between the groundtruth and the predicted bounding boxes. EAO can be seen as a combination of the robustness and the accuracy, which is the measure used to rank trackers in the VOT challenge. Comparison results between our tracker and the state-of-the-art methods are shown in Table~\ref{tab:VOT2018}. Our approach outperforms the previous best methods and achieves the SOTA results.

\begin{table}[t]
  \centering
  \begin{tabular}{l | c c c}
    \toprule
    Tracker & R & A & EAO \\
    \midrule
    DRT\cite{sun_correlation_2018} &  0.201 &  0.519 & 0.356 \\
    UPDT\cite{bhat_unveiling_2018} &  0.184 & 0.536 & 0.378 \\ 
    LADCF\cite{Xu2018LearningAD} & 0.159 & 0.503 & 0.389 \\
    ATOM\cite{danelljan_atom_2019} & 0.204 & 0.590 & 0.401 \\
    SiamRPN++\cite{li_siamrpn_2019} &  0.234 & 0.600 & 0.414 \\
    DiMP-50\cite{bhat_learning_2019} &  0.153 & 0.597 & 0.440 \\
    \midrule
    Ours-LT & 0.176 & 0.599 & \textcolor{blue}{0.441} \\
    Ours &  0.160 & 0.597 & \textcolor{red}{0.460} \\
    \bottomrule
  \end{tabular}
    \caption{State-of-the-art comparisons on the VOT2018 dataset in terms of robustness(R), accuracy(A) and expected average overlap (EAO). The best performing tracker is colored with \textcolor{red}{red}. The second best \textcolor{blue}{blue}. Our method achieves the best results among SOTA trackers. Ours-LT is our tracker with only long-term component.}
  \label{tab:VOT2018}
\end{table}

\textbf{TrackingNet}\cite{muller_trackingnet_2018}: The TrackingNet is a large scale dataset which consists of 511 test videos. The groundtruth of the test set are not given and predictions should be submitted to a public server for evaluation. The performance metrics are precision, normalized precision, and success. We show results in Table~\ref{tab:TrackingNet}. The DiMP tracker with ResNet-50 backbone achieves success 74.0\%. Our method uses the same backbone and achieves 75.5\%.
\begin{table}[t]
  \centering
  \begin{tabular}{l | c c c}
    \toprule
    Tracker & precision & normalized precision & success \\
    \midrule
    ECO\cite{danelljan_eco_2017} & 49.2 & 61.8 & 55.4 \\
    SiamFC\cite{bertinetto_fully-convolutional_2016} &  53.3 &  66.6 & 57.1 \\
    UPDT\cite{bhat_unveiling_2018} &  55.7 & 70.2 & 61.1 \\ 
    ATOM\cite{danelljan_atom_2019} & 64.8 & 77.1 & 70.3 \\
    SiamRPN++\cite{li_siamrpn_2019} &  69.4 & 80.0 & 73.3 \\
    DiMP-50\cite{bhat_learning_2019} &  68.7 & 80.1 & 74.0 \\
    \midrule
    Ours-LT & 68.6 & 79.8 & \textcolor{blue}{74.9} \\
    Ours &  69.6 & 80.7 & \textcolor{red}{75.5} \\
    \bottomrule
  \end{tabular}
    \caption{State-of-the-art comparisons on the TrackingNet dataset in terms of precision, normalized precision and success. Our method achieves the best results among SOTA trackers.}
  \label{tab:TrackingNet}
\end{table}

\begin{figure}[h]
  \centering
  \includegraphics[width=1.0\linewidth]{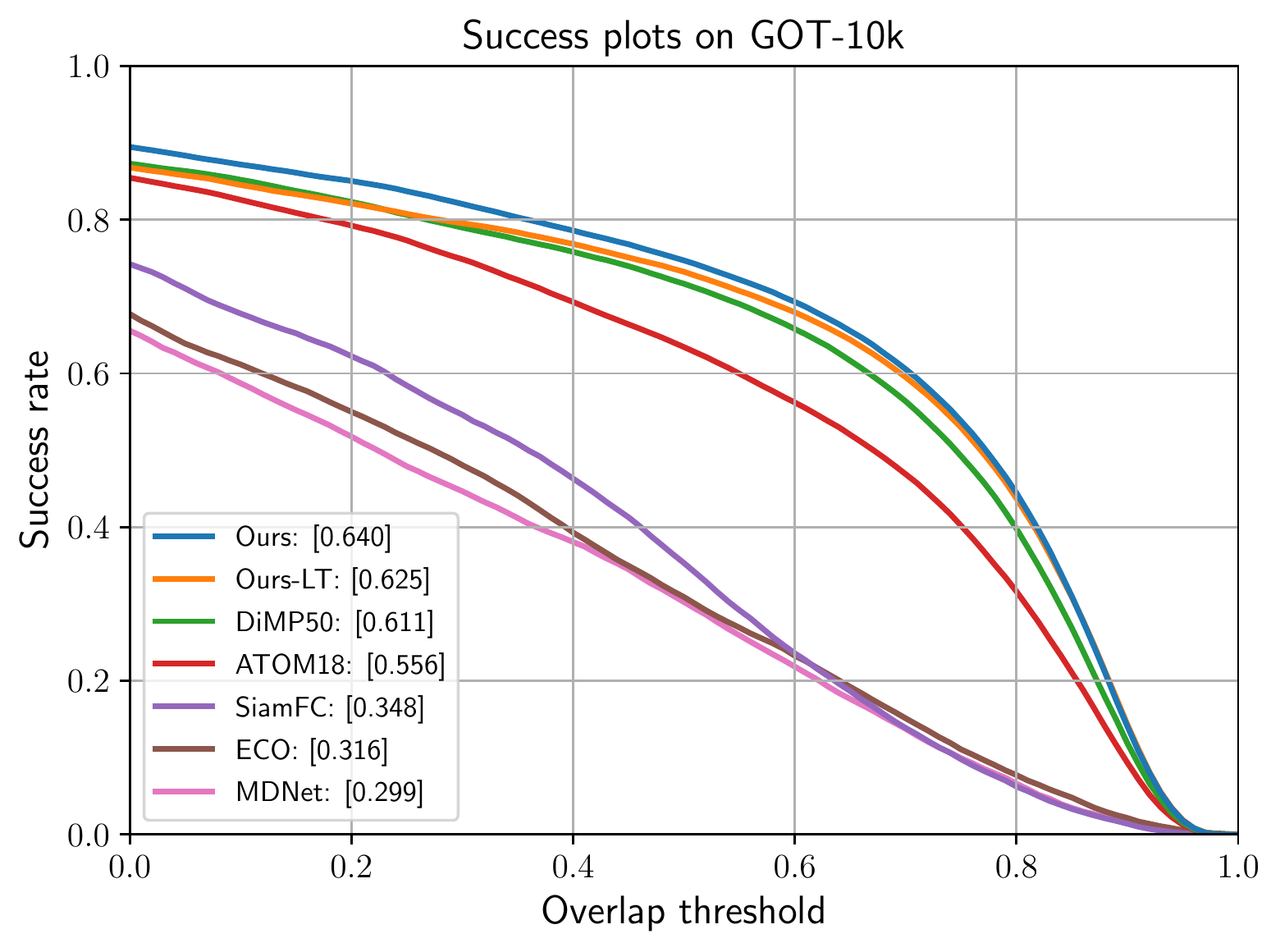}
  \caption{AUC curve on the GOT-10k datasets. Our tracker achieves the best result. \label{fig:success}}
\end{figure}

\textbf{GOT-10k}\cite{huang_got-10k_2019}: This is a large-scale dataset containing over 10,000 videos of which 180 videos are used for testing. This dataset is unique since there is no overlap in object categories between the training set and the test set. The model should be trained on the training set without using extra data. Therefore, tracking performance on unseen object categories is vital in this dataset. We follow the evaluation protocol and train all our models using only the training set of GOT-10k and submits the testing results to the public server for evaluation. The results are shown in Figure~\ref{fig:success}. Our method significantly improves the SOTA performance from 0.611 to 0.640.

\textbf{OTB-100}\cite{wu_object_2015}: The OTB-100 dataset contains 100 videos. The aspect ratio of the targets has a limited change in this dataset. Therefore, methods that have the aspect ratio constraint such as multi-scale search achieve better results. Our method performs on par with the state-of-the-art trackers. The AUC scores are shown in Table~\ref{tab:NFS}.

\textbf{LaSOT}\cite{fan_lasot_2019}: The LaSOT dataset contains 280 test videos with an average of 2500 frames per sequence. Compared to other datasets, the targets in LaSOT frequently undergo occlusion and out of the views, making it difficult for reliable short-term modeling. Nevertheless, our tracker is able to perform on par with the SOTA trackers. The AUC scores are shown in Table~\ref{tab:NFS}.

\textbf{NFS}\cite{kiani_galoogahi_need_2017}: The NFS dataset consists of 100 videos with high frame rate (240 FPS). We evaluate our approach on the downsampled 30 FPS version of this dataset. This dataset contains videos where the target is fast moving among similar distractors. This is challenging for existing trackers. However, with the proposed short-term model which models fast variations and the global structure consistency, our tracker significantly improves tracking performance. The AUC scores are shown in Table~\ref{tab:NFS}. Our approach outperforms the previous best methods and achieves the SOTA results.

\begin{table}[t]
  \centering
  \begin{tabular}{l | c c c}
    \toprule
    Tracker & OTB-100 & NFS & LaSOT \\
    \midrule
    MDNet\cite{nam_learning_2016} & 0.678 & 0.422 & 0.397 \\
    CCOT\cite{danelljan_beyond_2016} & 0.671 & 0.488 & - \\
    ECO\cite{danelljan_eco_2017} & 0.691 & 0.466 & 0.324 \\
    UPDT\cite{bhat_unveiling_2018} & \textcolor{red}{0.702} & 0.537 & - \\
    ATOM\cite{danelljan_atom_2019} & 0.663 & 0.584 & 0.514 \\
    SiamRPN++\cite{li_siamrpn_2019} & \textcolor{blue}{0.696} & - & 0.496 \\
    DiMP-50\cite{bhat_learning_2019} & 0.684 & \textcolor{blue}{0.619} & \textcolor{red}{0.568} \\
    \midrule
    Ours-LT & 0.682 & 0.614 & \textcolor{blue}{0.566} \\
    Ours & 0.686 & \textcolor{red}{0.629} & 0.565 \\
    \bottomrule
  \end{tabular}
    \caption{State-of-the-art comparisons on the OTB-100, NFS and LaSOT dataset in terms of AUC score. Due to the randomness of the tracker, evaluation is repeated 5 times and we report the average results. We use the same model and hyperparameters for all datasets.}
  \label{tab:NFS}
\end{table}

\subsection{Conclusion}
In this work, we propose a model that can efficiently and effectively learn the transient variations of the whole scene and thus can handle the challenging problems of fast variations and distractions of similar objects. The model is embedded in a long-short term framework to achieve the balance between stability and plasticity. As a result, the tracker can make use of the short-term model to handle fast variations and distractors as well as using the long-term model to avoid tracker drifting. We extensively verify the effectiveness of the proposed method on large scale benchmarks, which demonstrates state-of-the-art performance.

\bibliographystyle{ACM-Reference-Format}
\balance
\bibliography{submission}

\end{document}